\documentclass[11pt,a4paper]{article}
\usepackage[hyperref]{emnlp-ijcnlp-2019}
\usepackage{times}
\usepackage{latexsym}

\usepackage{url}

\aclfinalcopy 

\usepackage{verbatim}

\usepackage{graphicx}

\title{Red Dragon AI at TextGraphs 2019 Shared Task: \\
Language Model Assisted Explanation Generation}

\author{Yew Ken Chia \\
  Red Dragon AI  \\
  Singapore \\
  {\tt ken@reddragon.ai} \\\And
  Sam Witteveen \\
  Red Dragon AI  \\
  Singapore \\
  {\tt sam@reddragon.ai} \\\And
  Martin Andrews \\
  Red Dragon AI  \\
  Singapore \\
  {\tt martin@reddragon.ai} \\}

\date{}

\begin{document}
\maketitle
\begin{abstract}

The TextGraphs-13 Shared Task on Explanation Regeneration \cite{Jansen:19} asked participants 
to develop methods to reconstruct gold explanations for elementary science questions.
Red Dragon AI's entries used the language of the questions and explanation text directly,
rather than a constructing a separate graph-like representation.
Our leaderboard submission placed us 3$^{rd}$ in the competition, 
but we present here three methods of increasing sophistication, 
each of which scored successively higher on the test set after the competition close.

\end{abstract}

\section{Introduction}

The Explanation Regeneration shared task asked participants to develop methods 
to reconstruct gold explanations for elementary science questions \cite{Clark2018ThinkYH}, 
using a new corpus of gold explanations \cite{JANSEN18.81}
that provides supervision and instrumentation 
for this multi-hop inference task. 

Each explanation is represented as an ``explanation graph'', 
a set of atomic facts (between 1 and 16 per explanation, 
drawn from a knowledge base of 5,000 facts) that, together, 
form a detailed explanation for the reasoning required to answer and explain the resoning behind a question. 

Linking these facts to achieve strong performance at rebuilding the gold explanation graphs 
requires methods to perform multi-hop inference - which has been shown to be 
far harder than inference of smaller numbers of hops \cite{jansen2018multi}, 
particularly for the case here, where there is considerable uncertainty (at a lexical level)
of how individual explanations logically link somewhat `fuzzy' graph nodes.

\subsection{Dataset Review}

The WorldTree corpus \cite{JANSEN18.81} is a new dataset is 
a comprehensive collection of elementary science exam questions and explanations. 
Each explanation sentence is a fact that is related to science or common sense, 
and is represented in a structured table that can be converted to free-text. 
For each question, the gold explanations have lexical overlap (i.e. having common words), 
and are denoted as having a specific explanation role such as 
{\tt CENTRAL} (core concepts); 
{\tt GROUNDING} (linking core facts to the question); and
{\tt LEXICAL GLUE} (linking facts which may not have lexical overlap).

\begin{table}[t]
\label{results-base}
\begin{center}
\begin{small}

\vskip -0.1in
{
\renewcommand{\arraystretch}{1.4}
\begin{tabular}{lrrrr}
\hline
Data & Python & Scala & Python & Leaderboard  \\[-3pt]
split & Baseline & Baseline & Baseline$^{1e9}$  & Submission  \\
\hline
Train  & 0.0810 & & 0.2214 & 0.4216 \\
Dev    & 0.0544 & 0.2890 & 0.2140 & 0.4358 \\
Test   & & & & 0.4017 \\
\hline
\end{tabular}
}

\end{small}
\end{center}
\vskip -0.1in
\caption{Base MAP scoring - where the Python Baseline$^{1e9}$ 
is the same as the original Python Baseline, but with the {\tt evaluate.py} code
updated to assume missing explanations have rank of $10^9$
}
\end{table}

\subsection{Problem Review}

As described in the introduction, the general task being posed is one of multi-hop inference, 
where a number of `atomic fact' sentences must be combined to form a coherent chain of reasoning 
to solve the elementary science problem being posed.  

These explanatory facts must be retrieved from a semi-structured knowledge base - in which
the surface form of the explanation is represented as a series of terms gathered by their 
functional role in the explanation.  

For instance, for the explanation ``Grass snakes live in grass'' is encoded as ``[Grass snakes] [live in] [grass]'', 
and this explanation is found in a {\tt PROTO-HABITATS} table.  
However, in the same table there are also more elaborate explanations, for example :
``Mice live in in holes in the ground in fields / in forests.'' 
is expressed as : 
``[mice] [live in] [in holes in the ground] [in fields OR in forests]''.
And more logically complex :
``Most predators live in/near the same environment as their prey.'' 
being expressed as : 
``[most] [predators] [live in OR live near] [the same environment as their prey]''.

So, whereas the simpler explanations fit in the usual Knowledge-Base triples paradigm, 
the more complex ones are much more nuanced about what actually constitutes a node,
and how reliable the arcs are between them.  
Indeed, there is also a collection of {\tt if/then} explanations,
including examples such as :
``[if] [something] [has a] [positive impact on] [something else] 
[then] [increasing] [the] [amount of] [that something] [has a] [positive impact on] [that something else]'' - 
where the explanation has meta-effect on the graph itself, and includes `unbound variables'.
\footnote{The {\tt PROTO-IF-THEN} explanation table should have been annotated with a big red warning sign}

\section{Preliminary Steps}

In this work, we used the pure textual form of each explanation, problem and {\it correct} answer, 
rather than using the semi-structured form given in the column-oriented files provided in the dataset.
For each of these we performed Penn-Treebank tokenisation, 
followed by lemmatisation using the lemmatisation files provided with the dataset, 
and then stop-word removal.\footnote{PTB tokenisation and stopwords from the {\tt NLTK} package)}

Concerned by the low performance of the Python Baseline method (compared to the Scala Baseline, 
which seemed to operate using an algorithm of similar `strength'), 
we identified an issue in the organizer's evaluation script 
where predicted explanations that were missing {\it any} of the 
gold explanations were assigned a MAP score of zero.  
This dramatically penalised the Python Baseline, since it was restricted to only 
returning 10 lines of explanation.  It also effectively forces all submissions to include 
a ranking over all explanations - a simple fix (with the Python Baseline rescored in Table 1)
will be submitted via GitHub.  This should also make the upload/scoring process faster, since
only the top {\raise.17ex\hbox{$\scriptstyle\sim$}}1000 explanation lines meaningfully contribute to the rank scoring.

\section{Model Architectures}

Although more classic graph methods were initially attempted, 
along the lines of \citet{kwon2018controlling}, 
where the challenge of semantic drift in multi-hop inference was analysed
and the effectiveness of information extraction methods was demonstrated,
the following 3 methods (which now easily surpass the score of our competition submission) 
were ultimately pursued due to their simplicity/effectiveness.

\begin{table}[h]
\label{results-new}
\begin{center}
\begin{small}

\vskip 0.1in
{
\renewcommand{\arraystretch}{1.4}
\begin{tabular}{lrrr}
\hline
Data & Optimised & Iterated & BERT  \\
split & TF-IDF & TF-IDF & Re-ranking  \\
\hline
Train  & 0.4525 & 0.4827 & 0.6677 \\
Dev    & 0.4581 & 0.4966 & 0.5089 \\
Test   & 0.4274 & 0.4576 & 0.4771 \\
\hline
Time   & 0.02 & 46.97 & 92.96 \\
\end{tabular}
}

\end{small}
\end{center}
\vskip -0.1in
\caption{MAP scoring of new methods.  
The timings are in seconds for the whole dev-set, 
and the BERT Re-ranking figure includes 
the initial Iterated TF-IDF step.}
\end{table}

\subsection{Optimized TF-IDF}

As mentioned above, the original TF-IDF implementation 
of the provided Python baseline script did not predict a full ranking, 
and was penalized by the evaluation script. 
When this issue was remedied, its MAP score rose to 0.2140. 

However, there are three main steps that significantly improve the performance of this baseline:

\begin{enumerate}
\item The original question text included all the answer choices, 
only one of which was correct (while the others are distractors). 
Removing the distractors resulted in improvement;

\item The TF-IDF algorithm is very sensitive to keywords. 
Using the provided lemmatisation set and NLTK for tokenisation helped to align the different forms 
of the same keyword and reduce the vocabulary size needed;

\item Stopword removal gave us approximately 0.04 MAP improvement throughout - 
removing noise in the texts that was evidently `distracting' for TF-IDF.
\end{enumerate}

As shown in Table 2, these optimisation steps increased the 
Python Baseline score significantly, 
without introducing algorithmic complexity.

\subsection{Iterated TF-IDF}

While graph methods have shown to be effective for multi-hop question answering, 
the schema in the textgraphs dataset is unconventional (as illustrated earlier). 
To counter this, the previous TF-IDF method was extended to simulate jumps between explanations, 
inspired by graph methods, but without forming any actual graphs: 

\begin{enumerate}

\item TF-IDF vectors are pre-computed for all questions and explanation candidates; 

\item For each question, the closest explanation candidate by cosine proximity is selected, 
and their TF-IDF vectors are aggregated by a {\tt max} operation;

\item The next closest (unused) explanation is selected, 
and this process was then applied iteratively up to {\tt maxlen=128} times\footnote{
This {\tt maxlen} value was chosen to minimise computation time, noting that explanation ranks below approximately 100 have
negligible impact on the final score.}, 
with the current TF-IDF comparison vector progressively increasing in expressiveness.  
At each iteration, the current TF-IDF vector was down-scaled by an exponential factor of the length of the current explanation set, 
as this was found to increase development set results by up to +0.0344.

\end{enumerate}

By treating the TF-IDF vector as a representation of the current chain of reasoning, 
each successive iteration builds on the representation to accumulate a sequence of explanations. 

The algorithm outlined above was additionally enhanced by adding a weighting factor
to each successive explanation as it is added to the cumulative TF-IDF vector.  
Without this factor, the effectiveness was lower because the TF-IDF representation itself 
was prone to semantic drift away from the original question. 
Hence, each successive explanation’s weight was down-scaled, and this was shown 
to work well.\footnote{Full, replicable code is available on GitHub for all 3 methods described here, at \url{https://github.com/mdda/worldtree_corpus/tree/textgraphs}}


\subsection{BERT Re-ranking}

Large pretrained language models have been proven effective on a wide range of downstream tasks, 
including multi-hop question answering, such as in \citet{liu2019roberta} on the RACE dataset,
and \citet{xu2019multi} which showed that large finetuned language models can be beneficial 
for complex question answering domains (especially in a data-constrained context). 

Inspired by this, we decided to adapt BERT \cite{devlin2018bert} - 
a popular language model that has produced competitive results on a variety of NLP tasks - 
for the explanation generation task. 

For our `BERT Re-ranking' method, we attach a regression head to a BERT Language Model.  
This regression head is then trained to predict a relevance score for each pair of question and explanation candidate.
The approach is as follows : 

\begin{enumerate}

\item Calculate a TF-IDF relevance score for every tokenised explanation against the 
tokenised `[Problem] [CorrectAnswer] [Gold explanations]' {\it in the training set}.  
This will rate the true explanation sentences very highly, 
but also provide a `soft tail' of rankings across all explanations;

\item Use this relevance score as the prediction target of the BERT regression head, 
where BERT makes its predictions from the original 
`[Problem] [CorrectAnswer]' text combined with each potential Explanation text in turn (over the training set);

\item At prediction time, the explanations are ranked according 
to their relevance to `[Problem] [CorrectAnswer]' as predicted by the BERT model's output.
\end{enumerate}
 


We cast the problem as a regression task (rather than a classification task), since
treating it as a task to classify which explanations are relevant would result in an imbalanced dataset 
because the gold explanation sentences only comprise a small proportion of the total set. 
By using soft targets (given to us by the TF-IDF score against the gold answers in the training set), 
even explanations which are not designated as ``gold'' but have some relevance to the gold paragraph 
can provide learning signal for the model. 

Due to constraints in compute and time, the model is only used to rerank 
the $top_n=64$ predictions made by the TF-IDF methods. 

The BERT model selected was of ``Base'' size with 110M parameters, 
which had been pretrained on BooksCorpus and English Wikipedia.
We did not further finetune it on texts similar to the TextGraphs dataset prior to regression training. 
In other tests, we found that the ``Large'' size model did not help improve the final MAP score. 

\section{Discussion}


The authors' initial attempts at tackling the Shared Task focussed on graph-based methods.
However, as identified in \cite{jansen2018multi}, the uncertainty involved with interpreting each 
lexical representation, combined with the number of hops required, meant that this 
line of enquiry was put to one side\footnote{Having only achieved 0.3946 on the test set}.

While the graph-like approach is clearly attractive from a reasoning point of view 
(and will be the focus of future work), we found that using purely 
the textual aspects of the explanation database bore fruit more readily.
Also. the complexity of the resulting systems could be minimised 
such that the description of each system could be as consise as possible.

Specifically, we were able to optimise the TF-IDF baseline
to such an extent that our `Optimised TF-IDF' would now place $2^{nd}$ in the submission rankings, 
even though it used no special techniques at all.\footnote{Indeed, our Optimized TF-IDF, 
scoring 0.4581 on the dev set, and 0.4274 on the test set, could be considered a new baseline
for this corpus, given its simplicity.}

The Iterated TF-IDF method, while more algorithmically complex, also does not need any training 
on the data before it is used.  
This shows how effective traditional text processing methods can be, when used strategically.

\begin{figure}[t]
  \vskip -0.1in
  \begin{center}
    \centerline{\includegraphics[width=\columnwidth]{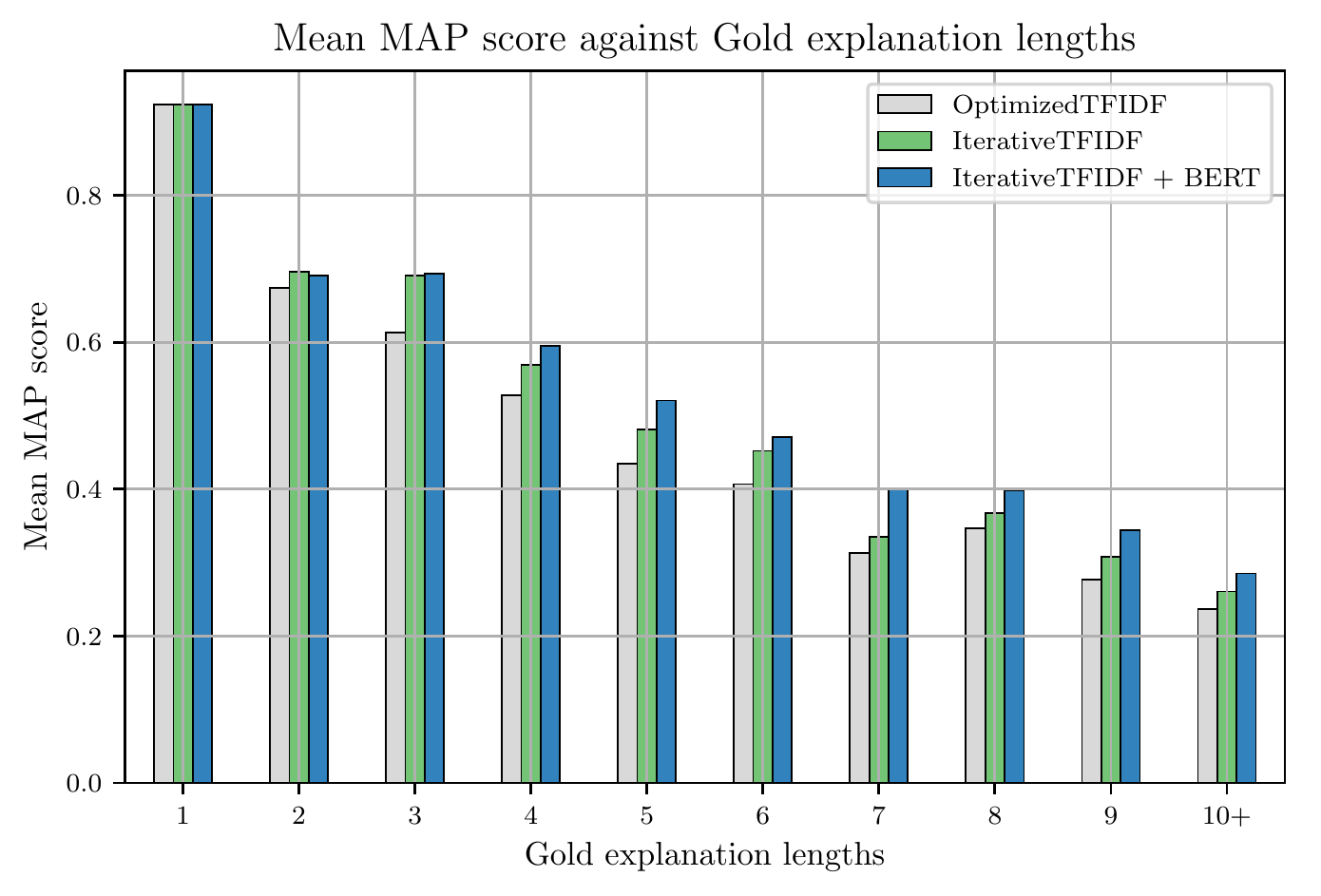}}
    \caption{
    Mean MAP score against gold explanation lengths
    }
    \label{map-vs-explanation-lengths}
  \end{center}
  \vskip -0.2in
\end{figure}

The BERT Re-ranking method, in contrast, does require training, and also applies one of the more
sophisticated Language Models available to extract more meaning from the explanation texts.

Figure 1 illustrates how there is a clear trend towards being able to 
build longer explanations as our semantic relevance methods become more sophisticated.

There are also clear trends across the data in Table 3 that show that the more sophisticated methods
are able to bring more {\tt CENTRAL} explanations into the mix, even though they are more
`textually distant' from the original Question and Answer statements.  
Surprisingly, this is at the expense of some of the {\tt GROUNDING} statements.

Since these methods seem to focus on different aspects of solving the ranking problem, 
we have also explored averaging the ranks they assign to the explanations 
(essentially ensembling their decisions).  
Empirically, this improves performance\footnote{The 
combination of `Iterated TF-IDF' and `BERT Re-ranking' scoring 0.5195 on the dev set, 
up from their scores of 0.4966 and 0.5089 respectively} 
at the expense of making the model more obscure.


\begin{table}[t]
\label{explanation-role}
\begin{center}
\begin{small}

\vskip 0.1in
{
\renewcommand{\arraystretch}{1.4}
\begin{tabular}{lrrr}
\hline
Explanation & Optimised & Iterated & BERT  \\
role & TF-IDF & TF-IDF & Re-ranking  \\
\hline
{\sc Grounding}  & 0.1373  & 0.1401  & 0.0880 \\
{\sc Lex-Glue}   & 0.0655  & 0.0733  & 0.0830 \\
{\sc Central}    & 0.4597  & 0.5033  & 0.5579 \\
\hline
{\sc Background} & 0.0302  & 0.0285  & 0.0349 \\
{\sc Neg}        & 0.0026  & 0.0025  & 0.0022 \\
{\sc Role}       & 0.0401  & 0.0391  & 0.0439 \\
\hline
\end{tabular}
}

\end{small}
\end{center}
\caption{Contribution of Explanation Roles - 
Dev-Set MAP per role (computed by filtering explanations of other roles out of 
the gold explanation list then computing the MAP as per normal)
}
\end{table}

\subsection{Further Work}

Despite our apparent success with less sophisticated methods, 
it seems clear that more explicit graph-based methods appears will be required to tackle the 
tougher questions in this dataset (for instance those that require logical deductions, 
as illustrated earlier, or hypothetical situations such as some `predictor-prey equilibrium' problems).
Even some simple statements (such as `Most predators ...') present
obstacles to existing Knowledge-Base representations.  

In terms of concrete next steps, we are exploring the idea of creating intermediate forms of representation, 
where textual explanations can be linked using a graph to plan out the logical steps.
However these grander schemes suffer from being incrementally less effective than
finding additional `smart tricks' for existing methods!

In preparation, we have begun to explore doing more careful preprocessing, notably :

\begin{enumerate}

\item Exploiting the structure of the explanation tables individually, since some
columns are known to be relationship-types that would be suitable for labelling arcs 
between nodes in a typical Knowledge Graph setting;

\item Expanding out the conjunction elements within the explanation tables.  
For instance in explanations like ``[coral] [lives in the] [ocean OR warm water]'',
the different sub-explanations ``(Coral, LIVES-IN, Ocean)'' and ``(Coral, LIVES-IN, WarmWater)''
can be generated, which are far closer to a `graph-able' representation;

\item Better lemmatisation : For instance `ice cube' covers both `ice' and `ice cube' nodes.
We need some more `common sense' to cover these cases.

\end{enumerate}

Clearly, it is early days for this kind of multi-hop inference over textual explanations.
At this point, we have only scratched the surface of the problem, and look forward
to helping to advance the state-of-the-art in the future.

\section*{Acknowledgments}
The authors would like to thank Google for access to the TFRC TPU program which 
was used in training and fine-tuning models during experimentation for this paper.

\bibliographystyle{acl_natbib}
\bibliography{../emnlp-ijcnlp-2019}

\begin{thebibliography}{8}
\expandafter\ifx\csname natexlab\endcsname\relax\def\natexlab#1{#1}\fi

\bibitem[{Clark et~al.(2018)Clark, Cowhey, Etzioni, Khot, Sabharwal, Schoenick,
  and Tafjord}]{Clark2018ThinkYH}
Peter~F. Clark, Isaac Cowhey, Oren Etzioni, Tushar Khot, Ashish Sabharwal,
  Carissa Schoenick, and Oyvind Tafjord. 2018.
\newblock \href {http://arxiv.org/abs/1803.05457} {Think you have solved
  question answering? {Try} {ARC}, the {AI2} {Reasoning} {Challenge}}.
\newblock \emph{ArXiv}, arXiv:1803.05457.

\bibitem[{Devlin et~al.(2018)Devlin, Chang, Lee, and
  Toutanova}]{devlin2018bert}
Jacob Devlin, Ming-Wei Chang, Kenton Lee, and Kristina Toutanova. 2018.
\newblock \href {http://arxiv.org/abs/1810.04805} {{BERT}: Pre-training of deep
  bidirectional transformers for language understanding}.
\newblock \emph{Computing Research Repository}, arXiv:1810.04805.

\bibitem[{Jansen(2018)}]{jansen2018multi}
Peter Jansen. 2018.
\newblock Multi-hop inference for sentence-level textgraphs: How challenging is
  meaningfully combining information for science question answering?
\newblock In \emph{Proceedings of the Twelfth Workshop on Graph-Based Methods
  for Natural Language Processing (TextGraphs-12)}, pages 12--17.

\bibitem[{Jansen and Ustalov(2019)}]{Jansen:19}
Peter Jansen and Dmitry Ustalov. 2019.
\newblock {TextGraphs 2019 Shared Task on Multi-Hop Inference for Explanation
  Regeneration}.
\newblock In \emph{Proceedings of the Thirteenth Workshop on Graph-Based
  Methods for Natural Language Processing (TextGraphs-13)}, Hong Kong.
  Association for Computational Linguistics.

\bibitem[{Jansen et~al.(2018)Jansen, Wainwright, Marmorstein, and
  Morrison}]{JANSEN18.81}
Peter Jansen, Elizabeth Wainwright, Steven Marmorstein, and Clayton Morrison.
  2018.
\newblock {WorldTree: A Corpus of Explanation Graphs for Elementary Science
  Questions supporting Multi-hop Inference}.
\newblock In \emph{Proceedings of the Eleventh International Conference on
  Language Resources and Evaluation (LREC 2018)}, Miyazaki, Japan. European
  Language Resources Association (ELRA).

\bibitem[{Kwon et~al.(2018)Kwon, Trivedi, Jansen, Surdeanu, and
  Balasubramanian}]{kwon2018controlling}
Heeyoung Kwon, Harsh Trivedi, Peter Jansen, Mihai Surdeanu, and Niranjan
  Balasubramanian. 2018.
\newblock Controlling information aggregation for complex question answering.
\newblock In \emph{European Conference on Information Retrieval}, pages
  750--757. Springer.

\bibitem[{Liu et~al.(2019)Liu, Ott, Goyal, Du, Joshi, Chen, Levy, Lewis,
  Zettlemoyer, and Stoyanov}]{liu2019roberta}
Yinhan Liu, Myle Ott, Naman Goyal, Jingfei Du, Mandar Joshi, Danqi Chen, Omer
  Levy, Mike Lewis, Luke Zettlemoyer, and Veselin Stoyanov. 2019.
\newblock Roberta: A robustly optimized bert pretraining approach.
\newblock \emph{arXiv preprint arXiv:1907.11692}.

\bibitem[{Xu et~al.(2019)Xu, Jansen, Martin, Xie, Yadav, Madabushi, Tafjord,
  and Clark}]{xu2019multi}
Dongfang Xu, Peter Jansen, Jaycie Martin, Zhengnan Xie, Vikas Yadav,
  Harish~Tayyar Madabushi, Oyvind Tafjord, and Peter Clark. 2019.
\newblock Multi-class hierarchical question classification for multiple choice
  science exams.
\newblock \emph{arXiv preprint arXiv:1908.05441}.

\end{thebibliography}


\end{document}